\documentclass[11pt]{article}

\usepackage[final]{acl}

\usepackage{array}
\usepackage{float}
\usepackage{times}
\usepackage{latexsym}
\usepackage{booktabs}
\usepackage{multirow} 
\usepackage{kotex}
\usepackage{placeins}
\usepackage{tabularx}
\usepackage{enumitem}

\usepackage[T1]{fontenc}

\usepackage[utf8]{inputenc}

\usepackage{microtype}

\usepackage{inconsolata}

\usepackage{graphicx}
\setkeys{Gin}{draft=false}
\usepackage{amsmath}
\usepackage{caption}

\title{Read the Room, Read the Image: Understanding Indirect Speech Acts in Multimodal Visual Contexts}

\author{
\textbf{Jaehee Kim\textsuperscript{1}\thanks{All authors contributed equally.}},
\textbf{Ji Hoon Chung\textsuperscript{1}\footnotemark[1]},
\textbf{Seoyoon Park\textsuperscript{2}\footnotemark[1]},
\textbf{Unsol Kim\textsuperscript{4}\footnotemark[1]},
\textbf{Kyungwon Park\textsuperscript{3}\footnotemark[1]},\\
\textbf{Ji Hak Kim\textsuperscript{2}\footnotemark[1]},
\textbf{Yi-Jun Chen\textsuperscript{1}\footnotemark[1]},
\textbf{Hansaem Kim\textsuperscript{2}\thanks{Corresponding author}}\\
\textsuperscript{1}Department of Korean Language and Literature, Yonsei University\\
\textsuperscript{2}Interdisciplinary Graduate Program of Linguistics and Informatics, Yonsei University\\
\textsuperscript{3}Department of Artificial Intelligence, Yonsei University; \textsuperscript{4}LG AI Research\\
\texttt{\{kim1016jh, chung.jihoon, seoyoon.park, jihakbabu,}\\
\texttt{cosmicboon, yijunchen, khss\}@yonsei.ac.kr},
\texttt{unsol.k12@gmail.com}
}

\begin{document}
\maketitle
\begin{abstract}
Indirect speech acts (ISAs) require pragmatic reasoning over context, since directive intent cannot be inferred from surface form alone. Prior text-based studies and multimodal benchmarks largely overlook this, focusing on explicitly encoded context or perceptual recognition particularly in high-context languages such as Korean. We introduce READI, a multimodal benchmark that evaluates ISA understanding through integrated reasoning over visual context and dialogue. Grounded in pragmatic theory, READI models graded indirectness and formulates the task as visual pragmatic question answering (V-PQA), enabling cross-lingual evaluation in English and Korean. Experiments show that even state-of-the-art multimodal models struggle with visually grounded ISAs with performance declining as indirectness increases, revealing fundamental limitations in interpreting indirectness. By proposing an evaluation paradigm for systematically assessing pragmatic understanding in multimodal settings, this study provides directions for improving the pragmatic reasoning abilities of language models.
\end{abstract}
\section{Introduction}
Linguistic meaning is not a fixed property inherent in linguistic form itself, but emerges through the relationship between an utterance and the situational context in which it is embedded. As argued by \citet{widdowson2004text}, linguistic expressions engage with the external world only when integrated with situational context, through which meaning is realized. From this perspective, language understanding should be viewed not as the interpretation of text in isolation, but as identifying how linguistic expressions are grounded in the real world through context. In line with this perspective, we define pragmatic understanding, following \citet{leech19836principles}, as the capacity to infer a speaker’s intended meaning beyond the literal content of an utterance, particularly by leveraging contextual and social cues.

Situational context is rarely encoded explicitly within the utterance itself; instead, it is provided as extra-linguistic factors arising through interaction. In real-world conversations, such contextual information is typically perceived through visual cognition within a physical environment. This dependence on extra-linguistic context becomes particularly salient in the interpretation of indirect speech acts, where speakers’ intentions are not directly expressed and must be inferred from contextual cues beyond linguistic form. For example, an utterance such as “It’s cold in here” may function not merely as a statement, but as an indirect directive to close a window, depending on the surrounding context.

In this study, we conceptualize pragmatic cues discussed in prior research~\citep{searle1975indirect,leech19836principles,brown1987politeness,hymes1974foundations} as sociopragmatic factors, and adopt the view that these factors play a central role in indirect speech act interpretation. From this perspective, we examine whether large language models (LLMs) can appropriately infer directive intentions embedded in indirect speech acts when sociopragmatic factors are presented visually.

Previous studies on intention inference and dialogue understanding in LLMs have primarily operationalized sociopragmatic factors as explicit textual input. While this approach has been effective for evaluating linguistic reasoning, it does not adequately reflect real conversational environments, in which contextual information is conveyed through non-verbal and physical means.

Recent multimodal LLM research has introduced evaluation frameworks based on Visual Question Answering (VQA), combining images and language, but existing benchmarks largely emphasize perceptual accuracy such as object or scene recognition. As LLMs increasingly function as interactive agents, they must move beyond recognizing visual content to interpreting images as resources encoding social interaction and sociopragmatic context. Motivated by this limitation, the present study aims to systematically evaluate how well LLMs can understand directive intentions in human indirect speech acts based solely on visually provided sociopragmatic factors.

To this end, we propose READI (Read the Room, Read the Image), a benchmark formulated as Visual Pragmatic Question Answering (V-PQA), which extends the traditional VQA paradigm by incorporating pragmatic inference. READI presents models with images depicting interactional situations alongside indirect directive utterances, requiring them to extract relevant contextual cues from the image to interpret the speaker’s intended directive meaning. Furthermore, by adopting a CCSARP (Cross-Cultural Speech Act Realization Project)-based graded indirectness design for both English and Korean, READI enables cross-linguistic comparison of indirect speech act understanding. Whereas existing multimodal LLM evaluations primarily emphasize image-level recognition accuracy, the present study focuses on a pragmatics-centered task grounded in contextual interpretation.

\begin{figure*}[t]
    \centering
    \includegraphics[width=\textwidth]{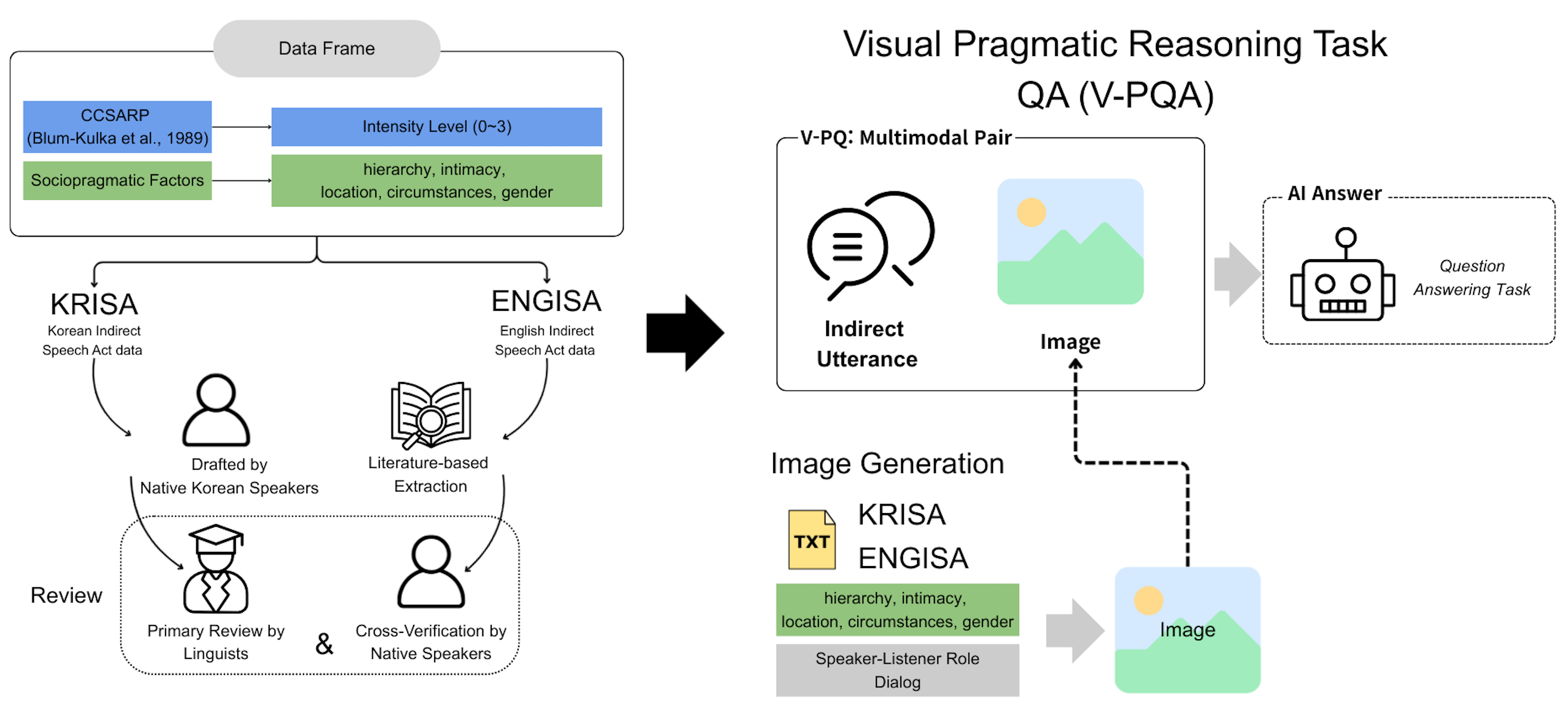}
    \caption{
    Overview of READI. The benchmark integrates indirect speech act data,
    sociopragmatic context, and visual information into a multimodal
    Visual Pragmatic Question Answering (V-PQA) task.
    }
    \label{fig:readi_overview}
\end{figure*}
\section{Related Work}

\subsection {Indirect Speech Act}
\textbf{Definition of ISAs.}
Indirect Speech Acts (ISAs) refer to utterances in which the speaker’s intended illocutionary force does not align with the utterance’s grammatical form or literal meaning, as originally theorized by \citet{searle1975indirect}.

ISAs are closely related to politeness theory. In Anglo-American contexts, indirectness often serves as a strategy for mitigating face-threatening acts \citep{brown1987politeness}, by reducing imposition on the addressee. However, cross-cultural studies suggest that this relationship is not universal \citep{ide199211, marti2006indirectness, yu2002culture, yu2011culture} and argue that in Japanese and Korean, deference is realized through interpersonal relations structured by social rank, status, and age, rather than through notions of non-imposition or option-giving. In such contexts, direct imposition may still be perceived as polite when framed with honorific or deferential forms.

Accordingly, \citet{blum1989investigating} proposed a cross-cultural framework for understanding ISAs through the Cross-Cultural Speech Act Realization Project (CCSARP). The CCSARP classifies directive speech acts by degree of indirectness into three levels: direct directives (DD), conventionally indirect directives (CID), and non-conventionally indirect directives (NCID). Direct directives explicitly encode the speaker’s intent; conventionally indirect directives rely on culturally conventionalized forms, whereas non-conventionally indirect directives require full pragmatic inference beyond linguistic form. The benchmark will provide a principled evaluation framework for indirect speech acts comprehension \citep{yum1988impact,kim1998high}.

\subsection {ISAs benchmark}

\textbf{Indirect Speech Acts in NLP.}
Previous research in natural language processing has primarily analyzed indirect speech acts (ISAs) in text-centered settings \citep{allen1980analyzing, wilske2006service, briggs2013hybrid}. More recent studies have begun to evaluate large language models’ (LLMs) ability to interpret ISAs using textualized contextual information. These works are meaningful in that they examine LLMs’ performance on indirect speech act understanding within text-based contexts. However, they typically operationalize sociopragmatic factors as explicitly specified text and focus on individual utterances rather than context-rich interactional settings.

Such configurations fail to adequately reflect real-world communication, which often relies on non-verbal and physical environmental cues. In particular, for non-conventionally indirect speech acts, the absence of sufficient context can lead to interpretive ambiguity even for human interlocutors. Moreover, prior findings indicate that LLMs are especially vulnerable when evaluating ISAs in high-context languages such as Korean \citep{koo2025evaluating}, and that the interpretation of non-conventionally indirect directives is highly dependent on model scale \citep{orsini2025direct}. Taken together, these results suggest that context-dependent indirect speech act inference remains a central challenge that has yet to be adequately addressed in NLP evaluation.

\textbf{Multimodal Benchmarks Limitations for ISAs.}
In multimodal research, Visual Question Answering (VQA)–style tasks have been widely used to combine images and language for reasoning\citep{agrawal2016analyzing, das2017visual,goyal2017making}. While such approaches demonstrate that visual information can be incorporated into language understanding, existing VQA-based benchmarks have primarily focused on object- and scene-level recognition. As a result, relatively little attention has been paid to how physical environments and interactional context contribute to utterance interpretation. In particular, such benchmarks typically rely on explicitly observable information and do not capture cases where meaning is pragmatically implied rather than directly stated, as is often the case in indirect speech acts.

Recognizing this limitation, \citet{zhang2025can} investigated whether models can interpret indirect speech acts in human–robot interaction (HRI) settings. Their study showed that identical utterances can activate different directive meanings depending on task context and shared situational conditions. However, the indirect expressions examined in this work are largely limited to conventionalized forms, and the evaluation does not systematically incorporate linguistically diverse levels of indirectness or richly structured visual context.

Overall, while prior multimodal studies demonstrate the feasibility of image–language reasoning, benchmarks that jointly consider indirect speech acts, graded linguistic indirectness, and visually grounded situational context remain scarce. This gap is particularly salient given that pragmatic context cannot be fully specified through text alone, but is often accessed through visual perception within physical and interactional environments.

\section{The READI Benchmark}

This section defines the READI benchmark tasks and describes the dataset construction process. 
READI evaluates whether LLMs can identify sociopragmatic factors depicted in images, use these cues to interpret the visual context, and infer the intended meaning of indirect utterances.\footnote{The dataset is publicly available at \url{https://github.com/jaeheehui/ISA-data}.}

\subsection{Dataset Construction}
\label{sec:dataset-construction}

\subsubsection{Indirect Speech Act (ISA) Dialogue Design}

This subsection describes the theoretical motivation and design procedure for the Korean and English dialogue data of indirect directive speech acts used in READI. READI evaluates higher-order pragmatic reasoning in visually grounded settings, where directive intent must be inferred from the interaction between linguistic expressions and situational context rather than from surface linguistic form alone.

A central design principle of READI is that the directive intent of an utterance cannot be reliably determined without access to the image-grounded situational context. Many target utterances are intentionally underspecified at the linguistic level and become interpretable as directives only when pragmatic cues embedded in the visual scene are taken into account. To operationalize indirectness in a cross-lingual and theoretically grounded manner, we adopt the Cross-Cultural Speech Act Realization Project (CCSARP) framework proposed by \citet{blum1989investigating}, which classifies directive speech acts according to indirectness levels.

We focus on conventionally indirect directives (CID) and non-conventionally indirect directives (NCID), further subdividing NCIDs according to pragmatic hint strength (strong vs. mild/no-hint). READI thus includes three indirectness levels-CID, NCID-Strong Hint, and NCID-Mild/No Hint-which become increasingly challenging as contextual inference becomes more critical. For details, see Appendix~\ref{sec:appendix-ccsarp}.

To ensure interpretation depends on contextual reasoning, READI controls five sociopragmatic factors: interlocutor relations, power asymmetry, social distance, interactional situation and physical setting, and the target and content of the directive act (Appendix ~\ref{sec:appendix-factors}). Importantly, these factors are modeled as interacting dimensions rather than reducing sociopragmatic context to a single variable such as power asymmetry, thereby avoiding an oversimplified interpretation of indirect speech acts. Dialogue scenarios instantiate diverse combinations of these factors, preventing over-reliance on lexical patterns and requiring pragmatic inference from situational configuration.

English (ENGISA) and Korean (KRISA) subsets were developed through language-specific procedures. ENGISA data were adapted from English-language pragmatics studies grounded in CCSARP and validated by seven native speakers and expert linguists. KRISA data were constructed in two stages: scenarios grounded in Korean interactional contexts were designed based on sociopragmatic factors, then native speakers drafted dialogues reviewed by linguists for pragmatic naturalness and label appropriateness.\footnote{Native speakers and domain experts were recruited via academic and professional networks. Informed consent was obtained from all participants. The data is used exclusively for research purposes, and all personal identifiers have been anonymized.}

\subsubsection{Multimodal Alignment and Image Generation}

In READI, images serve as pragmatic context for inferring directive intent in indirect utterances, rather than as targets for evaluating visual perception. Each image reflects the dialogue setting and target directive intent, including visual cues crucial for intent inference that cannot be derived from linguistic form alone.

To support cross-lingual comparison, we adopt a cultural persona alignment strategy, ensuring language-specific datasets depict characters aligned with their target sociocultural contexts. Specifically, KRISA images depict individuals and interpersonal relations typical of Korean society, whereas ENGISA images reflect sociopragmatic factors characteristic of English-speaking contexts. These visual cues are systematically designed to reflect the sociopragmatic factors defined in Section 3.1.1. In addition, during the image generation process, we intentionally incorporate controlled variation in demographic attributes such as age, gender, and social roles to avoid stereotypical associations between specific roles and identities, thereby enhancing diversity and reducing potential bias. All generated images are further reviewed by the authors to ensure alignment with intended sociopragmatic configurations and to mitigate unintended biases.

All images exclude textual elements or speech bubbles, ensuring the correct answer cannot be determined from linguistic information alone. With this design, READI evaluates pragmatic reasoning grounded in visual context, rather than visual perception or image recognition.

\subsection{Data Statistics}

The final READI benchmark is a multimodal dataset that combines images generated based on dialogue scenarios in Korean and English with utterances that realize indirect directive speech acts within the dialogues. Each READI task item consists of a single image, one indirect directive utterance, and four multiple-choice directive intent options. The final benchmark contains 102 multimodal items (57 Korean, 45 English). All items are drawn exclusively from indirect directive utterances at Levels 1–3, which inherently require pragmatic inference. The Korean subset was constructed by native Korean speakers residing in Korea, and the English subset was developed by native English speakers residing in the United States. To ensure sociocultural plausibility and reduce unintended demographic bias, participant roles and visual scenarios were designed with controlled variation in attributes such as age, gender, and social roles, followed by expert validation by linguists and pragmatics specialists.

The dataset size is designed to balance coverage of diverse sociopragmatic configurations with controlled experimental evaluation, enabling reliable comparison across models. In particular, each item reflects a distinct combination of contextual variables and indirectness levels, allowing the benchmark to probe generalizable pragmatic reasoning despite its moderate scale.

\begin{table}[t]
\centering
\small
\setlength{\tabcolsep}{3pt}

\begin{tabular}{l p{1.8cm} p{1.8cm}}
\toprule
\textbf{Category} & \textbf{Korean (KO)} & \textbf{English (EN)} \\
\midrule
Number of Items & 57 & 45 \\
\midrule
Target Levels 
& Lv1 (CID) & Lv1 (CID) \\
& Lv2, Lv3 (NCID) & Lv2, Lv3 (NCID) \\
\midrule
Min. Dialogue Turns & 3 & 3 \\
\midrule
Quality Control 
& 6 Experts & 7 Natives \\
& (4 Linguists) & \\
\midrule
Cultural Persona & Korean & Western/English \\
\bottomrule
\end{tabular}

\caption{READI Dataset Statistics Overview}
\label{tab:dataset_stats}

\end{table}

The dataset covers a wide range of domains, including offices, households, schools, and public spaces, and features diverse interpersonal relations such as supervisor–subordinate, colleagues, and shopkeeper–customer interactions. This diversity prevents overfitting to specific situations and encourages more generalized pragmatic reasoning.

\subsection{READI benchmark Task}
The READI benchmark is formulated as a four-choice multimodal Visual Question Answering (VQA) task designed to evaluate whether large language models can infer the intent of indirect directive speech acts by jointly interpreting an image $I$ and an utterance $U$. Each task item consists of an image $I$ encoding situational context, an indirect directive utterance $U$ whose intent must be inferred, and four answer options $C = \{c_1, c_2, c_3, c_4\}$. A multiple-choice setting is adopted to reduce open-ended ambiguity and enable controlled evaluation of fine-grained pragmatic distinctions.

The target of inference is a \textit{latent pragmatic meaning} $M$, which is defined conditionally on the situational and social context encoded in the image, rather than on the literal linguistic content of the utterance. In practice, $M$ is operationalized as the directive intent category represented by the correct answer option. Given the input $(I, U, C)$, a multimodal model $f_\theta$ estimates the probability that each option captures the intended meaning, and produces a prediction as follows:
\begin{equation*}
A = \arg\max_{c_i \in C} P_{\theta}(c_i \mid I, U)
\end{equation*}

The core challenge of the READI benchmark lies in requiring models to jointly integrate information from both the utterance $U$ and the image $I$. In many cases, the utterance is intentionally underspecified at the linguistic level, such that the intended directive meaning cannot be inferred from the utterance alone. Instead, models must rely on sociopragmatic cues embedded in the image—such as social relations, power dynamics, physical positioning, visual behaviors, and communicative goals—rather than on superficial lexical or structural similarities among the answer options.

To mitigate guessing based on surface-level cues, we apply a \textit{verb rotation strategy}, in which the underlying directive action is preserved while varying the directive verb across items (e.g., \textit{instruct}, \textit{command}, \textit{request}, \textit{suggest}). Distractor options are designed to include (i) \textit{literal traps} that reflect only the surface meaning of the utterance, and (ii) \textit{contextual mismatches} that are consistent with the visual scene but irrelevant to the speaker’s actual intent. All answer options are manually verified to ensure semantic plausibility and mutual exclusivity.

Additionally, models are required to explain which aspects of the image support their selected answer, providing a form of visual grounding for their predictions. This explanatory component is used only for qualitative analysis and is not included in the primary evaluation. Through this design, READI goes beyond text-only intent inference tasks and traditional VQA benchmarks by explicitly evaluating \textit{visually grounded sociopragmatic reasoning}.

Concrete examples of task instances and input--output formats are provided in the Appendix ~\ref{sec:appendix-2} for reproducibility and transparency.
\section{Experimental Settings}
\subsection{Models}
\begin{table}[t]
\centering
\small
\setlength{\tabcolsep}{4pt}
\begin{tabular}{lll}
\toprule
\textbf{Language} & \textbf{Commercial} & \textbf{Open-source} \\
\midrule
Multilingual
& GPT-5.2        & InternVL 3.5 (8B) \\
& Gemini 3       & LLaVA (7B)       \\
&                & Qwen3 (8B)       \\
\midrule
Korean-specific
& HyperCLOVA     & A.X-4.0-VL-Light (8B)         \\
&                & KoLLaVA (7B)     \\
\bottomrule
\end{tabular}
\caption{Model categorization by language coverage and licensing}
\label{tab:model_categories}
\end{table}
For the READI benchmark, we evaluated eight multimodal vision–language models (VLMs) selected based on two criteria: commercial versus open-source availability, and the presence of Korean-specific training or adaptation. These criteria allow us to examine how Korean linguistic and cultural grounding affects multimodal pragmatic reasoning while minimizing bias across model families. All open-source models were restricted to the 7–8B scale to ensure comparability (Table~\ref{tab:model_categories}).

The evaluated models fall into three groups: commercial general-purpose models (GPT-5.2, Gemini 3) as upper-bound references; open-source general-purpose models without Korean-specific training (InternVL 3.5, LLaVA, Qwen3); and Korean-specific multimodal models (HyperCLOVA, A.X-4.0-VL-Light, KoLLaVA). This grouping enables a focused comparison of whether Korean- and culture-specific training provides advantages in image-based indirect speech act intention inference. All models were evaluated in their released pretrained states.
\subsection{Evaluation Metrics}
READI is formulated as a multiple-choice classification task in which models infer a speaker’s indirect intention by interpreting a multimodal input consisting of an image and an utterance. Each item has a single correct, mutually exclusive answer, and the dataset is balanced across intention categories. Accordingly, we use accuracy as the primary evaluation metric.

Although F1 scores were also computed in preliminary experiments, they showed trends nearly identical to accuracy. To avoid redundant reporting and to ensure clear cross-model comparison, we therefore report accuracy only, which directly reflects whether a model correctly identifies the intended pragmatic category from the given multimodal context.

\section{Results}

\subsection{Quantitative Results}
\subsubsection{Overall Performance}
\begin{table*}[t]
\centering
\small
\setlength{\tabcolsep}{10pt}
\renewcommand{\arraystretch}{1.1}
\begin{tabular}{l l l cc cc}
\toprule
\multirow{2}{*}{Language} & \multirow{2}{*}{Model Type} & \multirow{2}{*}{Model}
& \multicolumn{2}{c}{\textbf{EN}} & \multicolumn{2}{c}{\textbf{KO}} \\
\cmidrule(lr){4-5} \cmidrule(lr){6-7}
& & & acc & macro\_f1 & acc & macro\_f1 \\
\midrule
multilingual & commercial & GPT-5.2            & 0.82 & 0.82 & 0.78 & 0.79 \\
multilingual & commercial & Gemini 3           & \textbf{0.93} & \textbf{0.94} & \textbf{0.88} &\textbf{0.88} \\
Korean       & commercial & HyperCLOVA         & 0.78 & 0.77 & 0.79 & 0.79 \\
\midrule
multilingual & open       & LLaVA              & 0.51 & 0.51 & 0.25 & 0.17 \\
Korean       & open       & KoLLaVA             & 0.29 & 0.22 & 0.33 & 0.25 \\
multilingual & open       & Qwen3               & 0.76 & 0.75 & 0.52 & 0.52 \\
multilingual & open       & InternVL 3.5   & 0.76 & 0.74 & 0.49 & 0.50 \\
Korean       & open       & A.X-4.0-VL-Light
                                              & 0.76 & 0.75 & 0.52 & 0.51 \\
\bottomrule
\end{tabular}
\caption{Overall performance on READI benchmark}
\label{tab:overall_performance}
\end{table*}

Table~\ref{tab:overall_performance} shows the overall performance of all models on the English and Korean datasets of the READI benchmark. Most models exhibit consistent performance degradation on the Korean dataset, indicating that understanding Korean indirect speech acts requires additional cultural context and pragmatic reasoning beyond basic multimodal integration. This pattern is observed across both commercial and open-source models, suggesting that the performance gap cannot be explained solely by model scale or training data size.

In contrast, Korean-specific models such as HyperCLOVA and KoLLaVA maintain or slightly improve their performance on Korean, highlighting the benefits of explicit language-specific training for multimodal pragmatic reasoning. Nevertheless, even the strongest models fall short of perfect accuracy, showing that image-based indirect speech act inference remains challenging in both languages, with limitations being more pronounced in Korean due to its stronger reliance on linguistic and cultural context.

\subsubsection{Performance Variation by ISA Intensity}
The preceding quantitative results revealed language-specific performance gaps in READI and the effects of Korean-specific training. However, overall accuracy alone does not fully explain which types of indirect speech acts pose greater challenges or which task characteristics drive performance degradation. To more precisely examine these limitations, we further analyze model accuracy as a function of indirectness levels.

\textbf{Language-wise Performance Patterns across Indirectness Levels} Because the degree of indirectness directly reflects the level of pragmatic reasoning required, READI annotates indirect speech acts with graded indirectness levels. Using these annotations, we investigate how model performance varies across different levels of indirectness, enabling a more fine-grained analysis of models’ pragmatic reasoning beyond aggregate accuracy (Table~\ref{tab:intensity_accuracy}).

\begin{table}[t]
\centering
\small
\renewcommand{\arraystretch}{0.9}  
\setlength{\tabcolsep}{8pt}        
\begin{tabular}{c cc}
\toprule
\textbf{Intensity} & \textbf{English (Acc)} & \textbf{Korean (Acc)} \\
\midrule
1 & 0.80 & 0.77 \\
2 & 0.85 & 0.58 \\
3 & \textbf{0.45} & 0.50 \\
\bottomrule
\end{tabular}
\caption{Average accuracy by indirectness levels}
\label{tab:intensity_accuracy}
\end{table}

First, when examining average accuracy across indirectness levels, the English dataset shows relatively similar performance at Lv1 and Lv2, followed by a sharp decline at Lv3. This pattern suggests that models can handle English indirect speech acts up to a moderate level of indirectness, but exhibit clear limitations in situational and pragmatic reasoning when explicit cues are largely absent.

In contrast, performance on the Korean dataset begins to degrade at Lv2, with most models showing substantial difficulty at Lv3. This indicates that Korean indirect speech acts require inference over social relations, contextual implicatures, and cultural expectations even at lower levels of indirectness.

\begin{table}[t]
\centering
\footnotesize
\setlength{\tabcolsep}{4pt}
\renewcommand{\arraystretch}{0.9}
\begin{tabular}{llcccc}
\toprule
\textbf{Language} & \textbf{Group} & \textbf{Lv1} & \textbf{Lv2} & \textbf{Lv3} & \textbf{Avg} \\
\midrule
\multirow{6}{*}{English}
& commercial (pooled)        & 0.91 & 0.98 & 0.64 & 0.84 \\
& ML-commercial         & 0.93 & 1.00 & 0.70 & \textbf{0.88} \\
& Korean-commercial          & 0.86 & 0.93 & 0.53 & 0.78 \\
& open (pooled)              & 0.73 & 0.77 & 0.33 & 0.61 \\
& ML-open               & 0.80 & 0.84 & 0.38 & 0.67 \\
& Korean-open                & 0.63 & 0.67 & 0.27 & 0.52 \\
\midrule
\multirow{6}{*}{Korean}
& commercial (pooled)        & 1.00 & 0.84 & 0.75 & 0.82 \\
& ML-commercial         & 1.00 & 0.86 & 0.75 & \textbf{0.83} \\
& Korean-commercial          & 1.00 & 0.81 & 0.76 & 0.80 \\
& open (pooled)              & 0.65 & 0.42 & 0.34 & 0.41 \\
& ML-open               & 0.78 & 0.41 & 0.33 & 0.40 \\
& Korean-open                & 0.50 & 0.43 & 0.36 & 0.41 \\
\bottomrule
\end{tabular}
\caption{Accuracy breakdown by language, model group, and indirectness intensity. ML indicates 'multilingual'.}
\label{tab:intensity_group_analysis}
\end{table}

\textbf{Effects of Language Adaptation and Model Type} To more systematically examine accuracy variations across indirectness levels, we aggregate models by language adaptation and by commercial versus open-source status, and report the results in Table~\ref{tab:intensity_group_analysis}. On the English dataset, multilingual commercial models achieve the highest overall performance and stability, whereas Korean-specific commercial models show relatively lower performance in English settings.

In contrast, a different pattern emerges on the Korean dataset. As indirectness intensity increases, Korean-specific commercial models consistently outperform multilingual commercial models, and among open-source models, Korean-adapted models exhibit smaller performance drops in high-intensity conditions. These results indicate that language adaptation contributes less to improving average accuracy than to mitigating performance degradation as pragmatic complexity increases.

Overall, the findings suggest that linguistic and cultural adaptation functions not merely as a source of performance gains, but as a key factor in ensuring robustness against increasing pragmatic demands.

\begin{table}[t]
\centering
\small
\begin{tabular}{lcc}
\toprule
\textbf{Language} & \textbf{Pearson $r$} & \textbf{Spearman $\rho$} \\
\midrule
English & \textbf{-0.45 (p = .002)} & \textbf{-0.35 (p = .017)} \\
Korean  & -0.22 (p = .109)         & -0.19 (p = .166)         \\
\bottomrule
\end{tabular}
\caption{Correlation between indirectness intensity and accuracy}
\label{tab:correlation_intensity}
\end{table}
\textbf{Correlation between Indirectness Level and Model Performance} As shown in Table~\ref{tab:correlation_intensity}, the Korean dataset reported no statistically significant correlation between indirectness level and model performance, indicating that performance degradation cannot be attributed solely to increasing task difficulty. Error analysis instead shows that models often fail to integrate sociopragmatic factors into culturally conventionalized speech act interpretations. These findings suggest that the primary bottleneck in understanding Korean indirect speech acts lies not in task difficulty itself, but in applying language- and culture-internalized pragmatic inference rules.

\begin{table}[t]
\centering
\small
\setlength{\tabcolsep}{3pt}          
\renewcommand{\arraystretch}{0.95}
\begin{tabular}{p{2.4cm}rrrr}        
\toprule
& \multicolumn{2}{c}{\textbf{English}} & \multicolumn{2}{c}{\textbf{Korean}} \\
\cmidrule(lr){2-3} \cmidrule(lr){4-5}
\textbf{Model} & Ablation & Baseline & Ablation & Baseline \\
\midrule
GPT-5.2               & 0.73 & \textbf{0.82} & 0.65 & \textbf{0.78} \\
Gemini 3             & 0.80 & \textbf{0.93} & 0.81 & \textbf{0.88} \\
LLaVA                & 0.47 & \textbf{0.51} & 0.27 & 0.25 \\
KoLLaVA              & 0.29 & 0.29           & 0.31 & \textbf{0.33} \\
Qwen3                & 0.69 & \textbf{0.76} & 0.50 & \textbf{0.52} \\
InternVL 3.5    & 0.64 & \textbf{0.76} & 0.42 & \textbf{0.49} \\
A.X-4.0-VL-Light  & 0.71 & \textbf{0.76} & 0.48 & \textbf{0.52} \\
\bottomrule
\end{tabular}
\caption{Ablation results with shuffled images compared to the baseline setting}
\label{tab:ablation_shuffle}
\end{table}

\subsubsection{Ablation Study: Image–Context Alignment}
To validate the legitimacy of READI, we conducted an ablation study to examine whether the task requirement of inferring speaker intention through image-based contextual interpretation is meaningfully reflected in model performance. Specifically, we randomly shuffled images across items so that each model received a visual context unrelated to the given utterance.

Results(Table~\ref{tab:ablation_shuffle}) show that, on both the English and Korean datasets, most models exhibit a clear drop in accuracy under the shuffled-image condition compared to the original setting. This indicates that models do not infer intentions solely from the utterance itself, but instead rely on the coherence between the utterance and its associated visual context. The observed performance degradation when images are decoupled from their corresponding utterances supports the claim that READI is not a purely text-based indirect speech act inference task, but a genuinely multimodal task requiring the integration of image-grounded contextual information.

These findings further demonstrate that READI does not include images merely as decorative elements, but rather constitutes an evaluation setting in which stable intention inference is difficult without reference to the specific visual context associated with each item.

\subsection{Qualitative Analysis}
\subsubsection{Error Analysis}
This section qualitatively examines the types of READI questions in which the models demonstrate relative strengths and those in which they show weaknesses. Overall, the models tend to infer speaker intent by relying on cases in which (i) the utterance includes conventional or explicit speech-act markers that quickly narrow the range of possible intent categories, or (ii) the sociopragmatic factors provided by the image relatively directly indicate the target action.

Meanwhile, a qualitative examination of the incorrect responses concentrated in the range where accuracy declines (Intensity Level Lv2–3) reveals that errors tend to increase as cues in the utterance weaken and the burden of integrative reasoning over sociopragmatic factors grows. The observed error types include (1) Contextual Miss and (2) Over/Under-interpretation.

Contextual Miss refers to errors in which models fail to connect the utterance with the sociopragmatic factors provided by the image and consequently do not recover an indirect directive intent as a request. Specifically, the majority of cases involved the models treating the utterance itself as a simple statement and failing to integrate scene cues provided by the image (e.g., a cluttered environment or a troubled facial expression) as sociopragmatic factors, leading them to select "situation description"-type options. Over-interpretation involves cases in which the model constructs unintended implicatures or misidentifies a simple utterance as an overly complex indirect speech act. In contrast, Under-interpretation occurs when the model fails to recover the intent of indirect speech acts and remains fixated on the literal meaning.

\subsubsection{Cultural Context Errors}
Cultural Context Errors refer to errors in which models misinterpret indirect directive intent as a non-directive utterance because they fail to apply the interactional norms conventionally expected in the relevant language community, even when they grasp the surface meaning of the utterance and scene information to some extent. This error type becomes particularly salient not simply as a translation or lexical difficulty, but when models fail to interpret the sociopragmatic factors jointly constructed by the utterance and the image as pragmatic conventions operative within that culture.

Cultural context errors were relatively rare in English; however, failures were observed in indirect expressions that are conventionally used in specific genres or situational contexts. In contrast, across models, Korean exhibited frequent cases of under-interpretation, indicating that the cultural characteristics underlying Korean indirect speech acts were not sufficiently reflected. The sources of these errors included limited understanding of Korean food culture, collectivist cultural norms, and generational differences. In addition, some errors were attributable to insufficient interpretation of cultural meanings embedded in ideophones and mimetic expressions. Most models failed to account for these cultural factors, and only in rare cases did a Korean-specialized model (HyperCLOVA) successfully recover the intended directive meaning.

\section{Discussion}
 While recent multimodal models excel in image-text alignment, they struggle significantly with understanding indirect speech act intentions. This degradation is more pronounced in Korean, confirming a heavy reliance on social and cultural contexts. Regarding performance changes by intensity, sharp drops were observed at specific levels rather than a gradual decline. English showed deterioration at the highest intensity, while Korean declined from mid-level intensity, as high-intensity acts require inferring both situation and intention without explicit cues.
 
Language-specific training proved beneficial in mitigating performance collapse in high-intensity ranges rather than simply improving average performance. Furthermore, unlike English, Korean performance could not be fully explained by intensity alone, suggesting a complex interplay of linguistic and pragmatic factors. The primary errors were contextual misses caused by failure to link sociopragmatic factors rather than object recognition failures. Consequently, multimodal evaluation must move beyond alignment to assess speech act function restoration, targeting cultural equivalence rather than translational isomorphism.
\section{Conclusion}
This study proposed READI, a multimodal benchmark for indirect speech acts, to analyze models' pragmatic inference capabilities stepwise. Our findings reveal that multimodal integration alone is insufficient for intent recovery and highlight significant performance gaps rooted in linguistic and cultural contexts. By providing a framework that integrates visual cues and pragmatic strategies, this research establishes a critical evaluation standard for AI to achieve a deeper understanding of complex human communicative contexts.

\section{Limitation}
Several limitations of the present study warrant discussion.
First, the analysis focuses exclusively on indirect directive speech acts. While directives provide a clear testing ground for indirectness and contextual dependence, other types of indirect speech acts are not considered, and the findings should therefore be interpreted within this scope.

Second, READI is based on theory-driven scenario design grounded in linguistic literature and expert review, rather than naturally occurring everyday conversations. Although this approach enables controlled manipulation of sociopragmatic variables, it may limit coverage of the spontaneity and variability characteristic of real-world interaction.

Third, the benchmark includes only two languages, English and Korean. While this allows for a focused comparison between a low-context and a high-context language, the results may not readily generalize to other linguistic and cultural settings.

Finally, the evaluation emphasizes models’ understanding of indirect speech acts, without assessing their ability to generate pragmatically appropriate responses. Further investigation is required to determine how multimodal pragmatic understanding relates to language production.


\section*{Acknowledgments}
 We are deeply grateful to the anonymous reviewers for their insightful comments and valuable suggestions, which have significantly contributed to enhancing the quality of this work. We also acknowledge that this research was supported by a grant from the Institute for AI and Social Innovation at Yonsei University (2025-22-0487).



\bibliography{custom}

@book{widdowson2004text,
  title={Text, context, pretext: Critical issues in discourse analysis},
  author={Widdowson, Henry G},
  year={2004},
  publisher={Oxford: Blackwell Publishing}
}

@article{agrawal2016analyzing,
  title={Analyzing the behavior of visual question answering models},
  author={Agrawal, Aishwarya and Batra, Dhruv and Parikh, Devi},
  journal={arXiv preprint arXiv:1606.07356},
  year={2016},
  url={https://arxiv.org/abs/1606.07356}
}

@article{blum1989investigating,
  title={Investigating cross-cultural pragmatics: An introductory overview},
  author={Blum-Kulka, Shoshana and House, Juliane and Kasper, Gabriele},
  journal={Cross-cultural pragmatics: Requests and apologies},
  volume={31},
  pages={1--34},
  year={1989}
}

@inproceedings{briggs2013hybrid,
  title={A hybrid architectural approach to understanding and appropriately generating indirect speech acts},
  author={Briggs, Gordon and Scheutz, Matthias},
  booktitle={Proceedings of the AAAI Conference on Artificial Intelligence},
  volume={27},
  number={1},
  pages={1213--1219},
  year={2013},
  url={https://ojs.aaai.org/index.php/AAAI/article/view/8537}
}

@book{brown1987politeness,
  title={Politeness: Some universals in language usage},
  author={Brown, Penelope and Levinson, Stephen C},
  volume={4},
  year={1987},
  publisher={Cambridge university press}
}

@inproceedings{das2017visual,
  title={Visual dialog},
  author={Das, Abhishek and Kottur, Satwik and Gupta, Khushi and Singh, Avi and Yadav, Deshraj and Moura, Jos{\'e} MF and Parikh, Devi and Batra, Dhruv},
  booktitle={Proceedings of the IEEE conference on computer vision and pattern recognition},
  pages={326--335},
  year={2017},
  url={https://arxiv.org/abs/1611.08669}
}

@inproceedings{goyal2017making,
  title={Making the v in vqa matter: Elevating the role of image understanding in visual question answering},
  author={Goyal, Yash and Khot, Tejas and Summers-Stay, Douglas and Batra, Dhruv and Parikh, Devi},
  booktitle={Proceedings of the IEEE conference on computer vision and pattern recognition},
  pages={6904--6913},
  year={2017},
  url={https://arxiv.org/abs/1612.00837}
}

@book{hymes1974foundations,
  title={Foundations in sociolinguistics: An ethnographic approach},
  author={Hymes, Dell},
  year={1974},
  publisher={Routledge}
}

@article{ide199211,
  title={11. The concept of politeness: An empirical study of American English and Japanese},
  author={Ide, Sachiko and Hill, Beverly and Carnes, Yukiko M and Ogino, Tsunao and Kawasaki, Akiko},
  journal={Politeness in language: Studies in its history, theory, and practice},
  pages={281},
  year={1992},
  doi={10.1515/9783110214080.281}
}

@article{kim1998high,
  title={High-versus low-context culture: A comparison of Chinese, Korean, and American cultures},
  author={Kim, Donghoon and Pan, Yigang and Park, Heung Soo},
  journal={Psychology \& Marketing},
  volume={15},
  number={6},
  pages={507--521},
  year={1998},
  doi={10.1002/(SICI)1520-6793(199809)15:6<507::AID-MAR2>3.0.CO;2-A}
}

@article{koo2025evaluating,
  title={Evaluating Large language models on Understanding Korean indirect Speech acts},
  author={Koo, Youngeun and Lee, Jiwoo and Park, Dojun and Park, Seohyun and Lee, Sungeun},
  journal={arXiv preprint arXiv:2502.10995},
  year={2025},
  url={https://arxiv.org/abs/2502.10995}
}

@book{leech19836principles,
  title={Principles of pragmatics},
  author={Leech, Geoffrey N},
  year={1983},
  publisher={Routledge}
}

@article{marti2006indirectness,
  title={Indirectness and politeness in Turkish--German bilingual and Turkish monolingual requests},
  author={Marti, Leyla},
  journal={Journal of Pragmatics},
  volume={38},
  number={11},
  pages={1836--1869},
  year={2006},
  doi={10.1016/j.pragma.2005.08.016}
}

@inproceedings{orsini2025direct,
  title={Direct and indirect interpretations of speech acts: evidence from human judgments and large language models},
  author={Orsini, Massimiliano and Brunato, Dominique},
  booktitle={Proceedings of the Eleventh Italian Conference on Computational Linguistics (CLiC-it 2025)},
  pages={837--848},
  year={2025},
  urt={https://aclanthology.org/2025.clicit-1.79/}
}

@incollection{searle1975indirect,
  title={Indirect speech acts},
  author={Searle, John R},
  booktitle={Speech acts},
  pages={59--82},
  year={1975},
  publisher={Brill}
}

@inproceedings{wilske2006service,
  title={Service robots dealing with indirect speech acts},
  author={Wilske, Sabrina and Kruijff, Geert-Jan},
  booktitle={2006 IEEE/RSJ International Conference on Intelligent Robots and Systems},
  pages={4698--4703},
  year={2006},
  doi={10.1109/IROS.2006.282502}
}

@article{yu2011culture,
  title={Culture-specific concepts of politeness: indirectness and politeness in English, Hebrew and Korean requests.},
  author={Yu, Kyong},
  journal={Intercultural pragmatics},
  volume={8},
  number={3},
  year={2011},
  doi={10.1515/iprg.2011.013}
}

@article{yu2002culture,
  title={Culture-Specific Concepts of Politeness: The Korean Concept of gongsonhada is different from the American English Concept of polite and from the Japanese Concept of teineina},
  author={Yu, Kyong-Ae},
  journal={Korean Journal of Applied Linguistics (in Korean)},
  volume={18},
  number={2},
  pages={41--60},
  year={2002}
}

@article{yum1988impact,
  title={The impact of Confucianism on interpersonal relationships and communication patterns in East Asia},
  author={Yum, June Ock},
  journal={Communications Monographs},
  volume={55},
  number={4},
  pages={374--388},
  year={1988},
  doi={10.1080/03637758809376178}
}

@inproceedings{zhang2025can,
  title={Can you pass that tool?: Implications of indirect speech in physical human-robot collaboration},
  author={Zhang, Yan and Ratnayake, Tharaka Sachintha and Sew, Cherie and Knibbe, Jarrod and Goncalves, Jorge and Johal, Wafa},
  booktitle={Proceedings of the 2025 CHI Conference on Human Factors in Computing Systems},
  pages={1--18},
  year={2025},
  url={https://dl.acm.org/doi/10.1145/3706598.3713780}
}

@article{allen1980analyzing,
  title={Analyzing intention in utterances},
  author={Allen, James F and Perrault, C Raymond},
  journal={Artificial intelligence},
  volume={15},
  number={3},
  pages={143--178},
  year={1980},
  doi={10.1016/0004-3702(80)90042-9}
}


\clearpage
\appendix
\onecolumn

\section{CCSARP Directness Levels}
\label{sec:appendix-ccsarp}


\vspace{-10pt}
\begin{table}[h]
\centering
\begin{tabularx}{\textwidth}{c l l X}
\toprule
\textbf{Level} & \textbf{Category} & \textbf{Description} & \textbf{Examples} \\
\midrule

-- & Direct & Explicit imperative & Open the window. \\

1 & CID & Conventionally indirect & Can you open the window? \\

2 & NCID-Strong hints & Indirect with lexical hint & I wish we had some fresh air. \\

3 & NCID-Mild hints & Indirect without explicit hint & The air feels stuffy. \\

\bottomrule
\end{tabularx}

\caption{(a) CCSARP Directness Levels with Descriptions and Examples}
\label{tab:ccsarp_full}
\end{table}

\vspace{-2pt}

\begin{table}[h]
\centering
\begin{tabularx}{\textwidth}{l X}
\toprule
\textbf{Category} & \textbf{Detailed Description} \\
\midrule

\textbf{DD} & The expression itself carries a directive meaning. \\
            & Interpretation as any other meaning is impossible. \\
            & Contextual inference is unnecessary. \\
            & The intended action or target is explicit. \\

\midrule

\textbf{CID} & The expression itself does not carry a directive meaning. \\
             & Uses expressions that are conventionally and frequently employed to convey a directive meaning. \\
             & The intended action or target is clearly revealed, but realized through a hedged (indirect) expression. \\
             & It can be interpreted as having meanings other than a directive. \\
             & Without contextual inference, it is impossible to conclude that the utterance functions as a directive among multiple possible interpretations. \\

\midrule

\textbf{NCID-Strong hints} & Contains no expressions that directly denote a directive meaning. \\
                          & Contains no expressions that are conventionally used to denote a directive meaning. \\
                          & Content related to the directive is present in the sentence (i.e., the expression provides a hint). \\
                          & Contextual inference is essential to identify the speaker's intent. \\
                          & The speaker's intent can be grasped through a single step of inference. \\

\midrule

\textbf{NCID-Mild hints} & Contains no expressions that directly denote a directive meaning. \\
                        & Contains no expressions that are conventionally used to denote a directive meaning. \\
                        & No content related to the directive appears in the sentence. \\
                        & Contextual inference is essential to identify the speaker's intent. \\
                        & Identifying the speaker's intent requires two or more steps of inference. \\

\bottomrule
\end{tabularx}

\caption{Detailed Annotation Guidelines and Inference Levels for ISA Data}
\label{tab:isa_guidelines}
\end{table}

\section{Sociopragmatic Factors}
\label{sec:appendix-factors}

\vspace{60pt}   

\noindent
\begin{tabular}{p{4cm} p{10cm}}
\toprule
\textbf{Category} & \textbf{Description} \\
\midrule

\textbf{(1) Interlocutor Relations} 
& Defines the participants in the dialogue with as many diverse roles as possible 
(e.g., shop owner--customer, doctor--patient, teacher--student, boss--employee, coworkers, customer--staff, physical trainer--member, superior--subordinate, etc.). \\
\midrule

\textbf{(2) Power Asymmetry \& Hierarchy} 
& Defines hierarchical relationships and social power structures based on authority and intimacy 
(e.g., superior--subordinate, senior--junior, manager--employee, teacher--student, customer--staff, junior--senior, subordinate--superior, physical trainer--member, doctor--patient, kitchen assistant--chef, grandparent--child, parent--child, adult--minor, etc.). \\
\midrule

\textbf{(3) Social Distance} 
& Indicates the level of intimacy in the relationship, categorized as high or low based on (1) and (2) 
(e.g., strangers, close friends, etc.). \\
\midrule

\textbf{(4) Interactional Situation and Physical Setting} 
& Specifies where the interaction takes place (e.g., home, workplace, public space, company, school, restaurant, café, airport, bus stop, hospital, department store, meeting room, parking lot, apartment, church, wedding hall, office, park, etc.). \\
\midrule

\textbf{(5) Target and Content of the Directive Act} 
& The ‘Target and Content of the Directive Act’ is included in all questions in the READI benchmark dataset. For example:

\textit{Question:} Considering the context of the image and the dialogue, what is the pragmatic function of the following statement when spoken by the person on the left? 
\newline \textit{Utterance:} ``Me too, I'm struggling here. I haven't been able to eat properly all day.'' \\
\bottomrule

\end{tabular}

\vspace{-6pt} 

\captionof{table}{Sociopragmatic factors of READI datasets}
\label{tab:sociopragmatic}
\clearpage

\section{Task Examples}
\label{sec:appendix-2}

\vspace{60pt}
\noindent
\begin{tabular}{p{3cm} p{10.5cm}}
\toprule
\multicolumn{2}{c}{\textbf{Example 1 (English)}} \\
\midrule

\textbf{ID} & ENGISA 6 \\

\textbf{Image} & \includegraphics[width=0.28\textwidth]{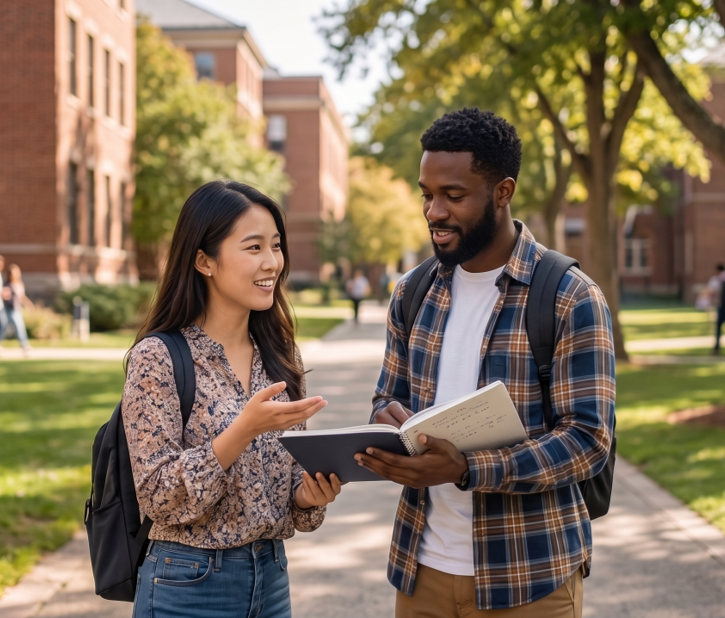} \\

\textbf{Question} &
\textit{Considering the context of the image and the dialogue, what is the pragmatic function of the following statement?} \\

\textbf{Utterance} &
\textit{“Oh, right! Do you think I could borrow your notes from yesterday’s class? Just for an hour so I can catch up.”} \\

\textbf{Choices} &
\textit{(1) To apologize for missing the class} \\
& \textit{(2) To borrow the notes so that she can catch up during the break} \\
& \textit{(3) To write something on the blackboard} \\
& \textit{(4) To explain why the class was missed} \\

\textbf{Answer} &
\textit{(2) To borrow the notes so that she can catch up during the break} \\

\textbf{ISA Intensity} & Lv1 (CID) \\

\bottomrule
\end{tabular}
\vspace{-6pt}
\captionof{table}{READI example in English.}
\label{tab:example-en}

\clearpage
\noindent
\begin{tabular}{p{3cm} p{10.5cm}}
\toprule
\multicolumn{2}{c}{\textbf{Example 2 (Korean)}} \\
\midrule

\textbf{ID} & KRISA 2 \\

\textbf{Image} & \includegraphics[width=0.28\textwidth]{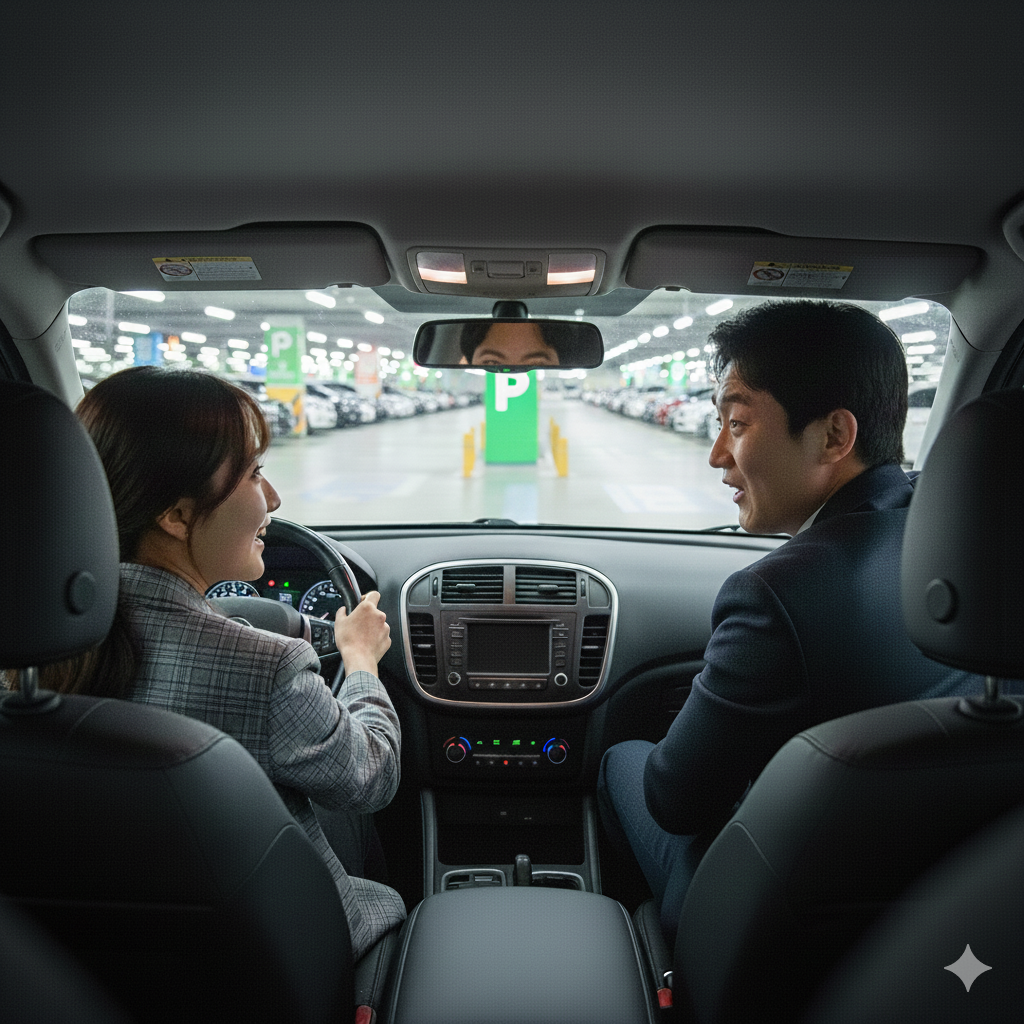} \\

\textbf{Question} &
이미지에서 오른쪽 사람이 다음과 같이 말하고 있다. 이미지를 고려했을 때 발화의 화용적 기능으로 알맞은 것은?\\
& \textit{In the image, the person on the right says the following. Considering the image, what is the pragmatic function of the utterance?} \\

\textbf{Utterance} &
“저기 빈 것 같은데?” \\
& \textit{“I think that spot over there is empty.”} \\

\textbf{Choices} &
(1) 저곳에 빈 자리가 있다는 설명 \\
& \textit{(1) A statement explaining that there is an empty space over there} \\
& (2) 빈 자리를 발견하지 못 하고 있는 운전자에 대한 질책 \\
& \textit{(2) A reproach toward a driver who has failed to notice the empty spot} \\
& (3) 뛰어가서 자리를 맡으라는 지시 \\
& \textit{(3) An instruction to run over and take the spot} \\
& (4) 빈 자리를 발견했으니 거기에 주차하는 것이 좋겠다는 제안 \\
& \textit{(4) A suggestion that, since the spot is empty, it would be good to park there} \\

\textbf{Answer} &
(4) 빈 자리를 발견했으니 거기에 주차하는 것이 좋겠다는 제안 \\
& \textit{(4) A suggestion that, since the spot is empty, it would be good to park there} \\

\textbf{ISA Intensity} & Lv2 (NCID-Strong hint) \\

\bottomrule
\end{tabular}

\vspace{-6pt}
\captionof{table}{READI example in Korean.}
\label{tab:example-ko}

\vspace{4pt}

\clearpage
\section{Detailed Performance Metrics for English and Korean READI}
\label{sec:appendix-detailed-results}
In this section, we provide a comprehensive breakdown of the experimental results for both English (Table \ref{tab:results-english-detailed}) and Korean (Table \ref{tab:results-korean-detailed}) subsets of the READI benchmark. The tables report Accuracy, Precision, Recall, and F1-score across three levels of indirectness and the overall average. 
The graded indirectness levels follow the CCSARP framework, where Level 1 represents conventionally indirect directives, and Levels 2 and 3 represent non-conventionally indirect directives with varying degrees of contextual hints. These detailed metrics further illustrate the performance gap between commercial and open-source models, as well as the impact of cultural and linguistic grounding on pragmatic reasoning.

\begin{center}
    \small
    \begin{tabular}{llcccc}
        \toprule
        \multicolumn{6}{c}{\textbf{Detailed Metrics for English ISA (READI-EN)}} \\
        \toprule
        \textbf{Model} & \textbf{Level} & \textbf{Accuracy} & \textbf{Precision} & \textbf{Recall} & \textbf{F1-score} \\
        \midrule
        \multirow{4}{*}{\textbf{GPT-5.2}} & 1 & 0.93 & 0.96 & 0.92 & 0.93 \\
        & 2 & 1.00 & 1.00 & 1.00 & 1.00 \\
        & 3 & 0.53 & 0.60 & 0.61 & 0.53 \\
        & \textbf{Avg} & \textbf{0.82} & 0.83 & 0.84 & \textbf{0.82} \\
        \midrule
        \multirow{4}{*}{\textbf{Gemini 3}} & 1 & 0.93 & 0.96 & 0.92 & 0.93 \\
        & 2 & 1.00 & 1.00 & 1.00 & 1.00 \\
        & 3 & 0.87 & 0.88 & 0.93 & 0.88 \\
        & \textbf{Avg} & \textbf{0.93} & 0.94 & 0.94 & \textbf{0.94} \\
        \midrule
        \multirow{4}{*}{\textbf{HyperCLOVA}} & 1 & 0.87 & 0.93 & 0.83 & 0.86 \\
        & 2 & 0.93 & 0.96 & 0.94 & 0.94 \\
        & 3 & 0.53 & 0.57 & 0.54 & 0.51 \\
        & \textbf{Avg} & \textbf{0.78} & 0.77 & 0.79 & \textbf{0.77} \\
        \midrule
        \multirow{4}{*}{\textbf{LLaVA}} & 1 & 0.53 & 0.61 & 0.56 & 0.52 \\
        & 2 & 0.73 & 0.89 & 0.71 & 0.72 \\
        & 3 & 0.27 & 0.43 & 0.35 & 0.27 \\
        & \textbf{Avg} & \textbf{0.51} & 0.61 & 0.55 & \textbf{0.51} \\
        \midrule
        \multirow{4}{*}{\textbf{KoLLaVA}} & 1 & 0.33 & 0.36 & 0.36 & 0.33 \\
        & 2 & 0.40 & 0.39 & 0.33 & 0.30 \\
        & 3 & 0.13 & 0.25 & 0.08 & 0.10 \\
        & \textbf{Avg} & \textbf{0.29} & 0.24 & 0.25 & \textbf{0.22} \\
        \midrule
        \multirow{4}{*}{\textbf{Qwen3}} & 1 & 0.93 & 0.96 & 0.92 & 0.93 \\
        & 2 & 0.93 & 0.96 & 0.94 & 0.94 \\
        & 3 & 0.40 & 0.40 & 0.36 & 0.37 \\
        & \textbf{Avg} & \textbf{0.76} & 0.75 & 0.76 & \textbf{0.75} \\
        \midrule
        \multirow{4}{*}{\textbf{InternVL 3.5}} & 1 & 0.93 & 0.96 & 0.92 & 0.93 \\
        & 2 & 0.87 & 0.90 & 0.85 & 0.86 \\
        & 3 & 0.47 & 0.53 & 0.48 & 0.45 \\
        & \textbf{Avg} & \textbf{0.76} & 0.75 & 0.76 & \textbf{0.74} \\
        \midrule
        \multirow{4}{*}{\textbf{A.X-4.0-VL-Light}} & 1 & 0.93 & 0.96 & 0.92 & 0.93 \\
        & 2 & 0.93 & 0.94 & 0.94 & 0.93 \\
        & 3 & 0.40 & 0.40 & 0.36 & 0.37 \\
        & \textbf{Avg} & \textbf{0.76} & 0.74 & 0.76 & \textbf{0.75} \\
        \bottomrule
    \end{tabular}
    \captionof{table}{Detailed performance metrics for the English READI benchmark (two decimal places).}
    \label{tab:results-english-detailed}
\end{center}

\vspace{2em} 

\begin{center}
    \small
    \begin{tabular}{llcccc}
        \toprule
        \multicolumn{6}{c}{\textbf{Detailed Metrics for Korean ISA (READI-KO)}} \\
        \toprule
        \textbf{Model} & \textbf{Level} & \textbf{Accuracy} & \textbf{Precision} & \textbf{Recall} & \textbf{F1-score} \\
        \midrule
        \multirow{4}{*}{\textbf{GPT-5.2}} & 1 & 1.00 & 1.00 & 1.00 & 1.00 \\
        & 2 & 0.82 & 0.81 & 0.84 & 0.80 \\
        & 3 & 0.67 & 0.66 & 0.67 & 0.65 \\
        & \textbf{Avg} & \textbf{0.78} & 0.77 & 0.78 & \textbf{0.79} \\
        \midrule
        \multirow{4}{*}{\textbf{Gemini 3}} & 1 & 1.00 & 1.00 & 1.00 & 1.00 \\
        & 2 & 0.89 & 0.89 & 0.93 & 0.89 \\
        & 3 & 0.83 & 0.82 & 0.83 & 0.81 \\
        & \textbf{Avg} & \textbf{0.88} & 0.87 & 0.88 & \textbf{0.88} \\
        \midrule
        \multirow{4}{*}{\textbf{HyperCLOVA}} & 1 & 1.00 & 1.00 & 1.00 & 1.00 \\
        & 2 & 0.81 & 0.81 & 0.81 & 0.81 \\
        & 3 & 0.76 & 0.72 & 0.73 & 0.72 \\
        & \textbf{Avg} & \textbf{0.79} & 0.80 & 0.80 & \textbf{0.79} \\
        \midrule
        \multirow{4}{*}{\textbf{LLaVA}} & 1 & 0.33 & 0.11 & 0.33 & 0.17 \\
        & 2 & 0.31 & 0.13 & 0.23 & 0.16 \\
        & 3 & 0.11 & 0.08 & 0.17 & 0.10 \\
        & \textbf{Avg} & \textbf{0.25} & 0.11 & 0.22 & \textbf{0.17} \\
        \midrule
        \multirow{4}{*}{\textbf{KoLLaVA}} & 1 & 0.20 & 0.06 & 0.25 & 0.10 \\
        & 2 & 0.34 & 0.27 & 0.30 & 0.28 \\
        & 3 & 0.33 & 0.26 & 0.30 & 0.27 \\
        & \textbf{Avg} & \textbf{0.33} & 0.25 & 0.27 & \textbf{0.25} \\
        \midrule
        \multirow{4}{*}{\textbf{Qwen3}} & 1 & 1.00 & 1.00 & 1.00 & 1.00 \\
        & 2 & 0.48 & 0.49 & 0.50 & 0.49 \\
        & 3 & 0.44 & 0.46 & 0.50 & 0.43 \\
        & \textbf{Avg} & \textbf{0.52} & 0.50 & 0.51 & \textbf{0.52} \\
        \midrule
        \multirow{4}{*}{\textbf{InternVL 3.5}} & 1 & 1.00 & 1.00 & 1.00 & 1.00 \\
        & 2 & 0.43 & 0.47 & 0.47 & 0.44 \\
        & 3 & 0.44 & 0.47 & 0.46 & 0.43 \\
        & \textbf{Avg} & \textbf{0.49} & 0.50 & 0.47 & \textbf{0.50} \\
        \midrule
        \multirow{4}{*}{\textbf{A.X-4.0-VL-Light}} & 1 & 1.00 & 1.00 & 1.00 & 1.00 \\
        & 2 & 0.52 & 0.49 & 0.58 & 0.51 \\
        & 3 & 0.39 & 0.40 & 0.38 & 0.38 \\
        & \textbf{Avg} & \textbf{0.52} & 0.49 & 0.52 & \textbf{0.51} \\
        \bottomrule
    \end{tabular}
    \captionof{table}{Detailed performance metrics for the Korean READI benchmark (two decimal places).}
    \label{tab:results-korean-detailed}
\end{center}

\clearpage

\end{document}